\documentclass[10pt,twocolumn,letterpaper]{article}

\usepackage[pagenumbers]{cvpr} 

\usepackage[accsupp]{axessibility}  %
\usepackage{graphicx}
\usepackage{amsmath}
\usepackage{amssymb}
\usepackage{booktabs}
\usepackage{bm}
\usepackage{dsfont}
\usepackage{enumerate}
\usepackage{multirow}
\usepackage{makecell}
\usepackage[pagebackref,breaklinks,colorlinks]{hyperref}

\usepackage[capitalize]{cleveref}
\crefname{section}{Sec.}{Secs.}
\Crefname{section}{Section}{Sections}
\Crefname{table}{Table}{Tables}
\crefname{table}{Tab.}{Tabs.}

\def\cvprPaperID{*****} 
\def\confName{CVPR}
\def\confYear{2022}

\begin{document}

\title{Edge-enhanced Feature Distillation Network for Efficient Super-Resolution}
\author{Yan Wang\\
Nankai-Baidu Joint Lab, Nankai University\\
{\tt\small wyrmy@foxmail.com}}

\maketitle

\begin{abstract}
With the recently massive development in convolution neural networks, numerous lightweight CNN-based image super-resolution methods have been proposed for practical deployments on edge devices. However, most existing methods focus on one specific aspect: network or loss design, which leads to the difficulty of minimizing the model size. To address the issue, we conclude block devising, architecture searching, and loss design to obtain a more efficient SR structure. In this paper, we proposed an edge-enhanced feature distillation network, named EFDN, to preserve the high-frequency information under constrained resources. In detail, we build an edge-enhanced convolution block based on the existing reparameterization methods. Meanwhile, we propose edge-enhanced gradient loss to calibrate the reparameterized path training. Experimental results show that our edge-enhanced strategies preserve the edge and significantly improve the final restoration quality. Code is available at \url{https://github.com/icandle/EFDN}.
 
\end{abstract}

\section{Introduction}
\label{sec:intro}

\begin{figure}[!t]
  \setlength\tabcolsep{4.0pt} 
  \centering
  \small
  \begin{tabular}{ccc}
    \multirow{-6.5}{*}{\includegraphics[width=.64\linewidth, height=5.3cm]{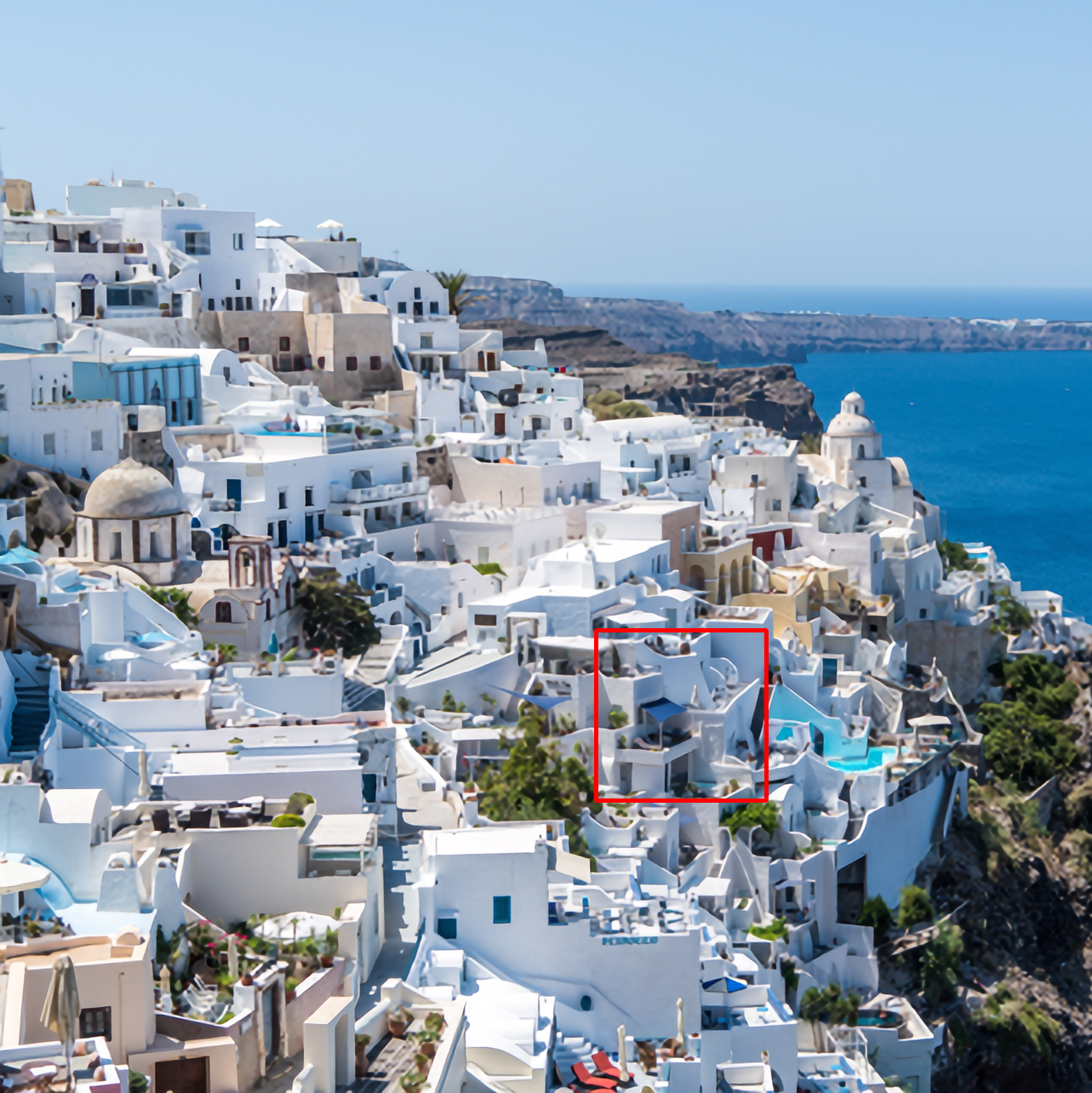}} 
    & \includegraphics[width=.3\linewidth, height=2.4cm]{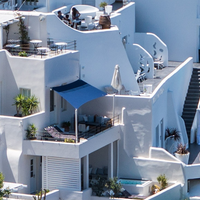} \\
    & HR\\
    & \includegraphics[width=.3\linewidth, height=2.4cm]{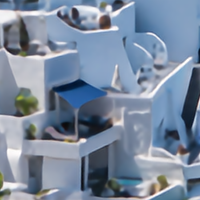} \\
    \emph{0823} from DIV2K & SR\\
  \end{tabular}\\
  \setlength\tabcolsep{4.0pt} 
  \begin{tabular}{ccc}
  \includegraphics[width=.3\linewidth, height=2.4cm]{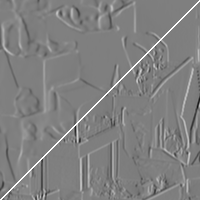}&\includegraphics[width=.3\linewidth, height=2.4cm]{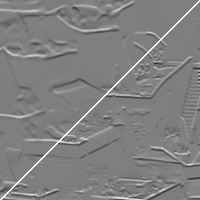}& \includegraphics[width=.3\linewidth, height=2.4cm]{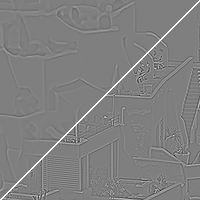} \\
  $\mathcal{L}_{x}=0.0201$& $\mathcal{L}_{y}=0.0092$ &$\mathcal{L}_{l}=0.0015$
 
\end{tabular}
  \caption{Visual results on DIV2K~\cite{div2k}. The lower right parts are ground truth gradient maps processed by the Sobel-(x,y) and Laplacian filters. The upper left part are results for EFDN.  }
  \label{fig:div2k}
  \end{figure}

Image super-resolution (SR) is a widely concerning low-level computer vision task that aims to build missing high-frequency information in degraded low-quality images. However, it is challenging to predict the appropriate images due to the ill-posed nature since one high-resolution (HR) image corresponds to plenty of low-resolution images. Many classical methods~\cite{Edge-interp,ANR,A+} have been proposed to address this problem, but their reconstruction quality under large magnification is often unsatisfactory. Recently, many convolution neural network (CNN) based approaches~\cite{SRCNN,FSRCNN,VDSR,EDSR,RCAN,li2019learning,HAN} were introduced to obtain realistic super-resolution images. Notwithstanding, most of them focus on improving restoration quality while involuntarily increasing the model scale, resulting in difficulty for mobile devices deployment. In this paper, we mainly concentrate on deploying SR models under resource-limited conditions.

In order to design a qualified lightweight neural network, researchers cut into this problem from the perspective of parameter reduction and calculation reduction. Sticking to reducing the size of convolution and features, FSRCNN~\cite{FSRCNN} first employed the post upscaling module, which removed both calculations and parameters. For more significant parameter reduction, the recurrent learning is leveraged in many works, including DRCN~\cite{DRCN} and DRRN~\cite{DRRN}. However, these recursive approaches cost more computation resources due to their limited representation capability. For instance, 17.9T multiply-add operations (MAdds) are spent in DRCN and 6.8T in DRRN, which are unbearable for mobile devices. Therefore, the researchers have shifted the critical point of efficient SR to designing effective modules and dedicated networks.

To this end, networks~\cite{CARN, IDN, IMDN} and blocks~\cite{ECB,DBB,ACB,PAN} were proposed to improve efficiency. In RFDN~\cite{RFDN}, the shallow residual block and dedicated distillation procedure are introduced to achieve superior performance under constrained conditions. However, its parameters and operations are still unaffordable for part of edge devices due to the dense connections. Zhang \etal~\cite{ECB} utilized a plain network with a re-parameterizable convolution block to accomplish real-time application on commodity mobile devices. But they suffer from PSNR drop like other simple networks. In addition to these manually designed networks and blocks, many methods based on neural architecture search (NAS)~\cite{ song2020efficient,wu2021trilevel,chu2021fast} and model pruning~\cite{li2020dhp,li2021heterogeneity} are proposed to obtain more flexible and lightweight networks. DLSR~\cite{DLSR} introduced a differentiable NAS method to find a more flexible typology based on RFDN, which designs cell-level and network-level with meager cost. Nevertheless, DLSR overuses depth-wise convolution, bringing a large amount of activation and memory consumption. Hence, there is still room for improvement in designing an efficient SR model.

In order to address the above issues, we propose an efficient \underline{E}dge-enhanced \underline{F}eature \underline{D}istillation \underline{N}etwork (EFDN), which combines block composing, architecture searching, and loss designing to obtain a trade-off between the performance and light-weighting. For block composing, we sum up the re-parameterization methods~\cite{ECB,DBB,ACB,RepVGG} and design a more effective and complex edge-enhanced diverse branch block. In detail, we employ several reasonable re-parameterizable branches to enhance the structural information extraction, and then we integrate them into a vanilla convolution to maintain the inference performance. To ensure the effective optimization of parallel branches in EDBB, we design an edge-enhanced gradient-variance loss (EG) based on gradient-variance loss~\cite{GV}. The proposed loss enforces minimizing the difference between the computed variance maps, which is helpful to restore sharper edges. As shown in \cref{fig:div2k}, we present the gradient maps calculated by different filters and the corresponding EG loss. In addition, the NAS strategy of DLSR is adopted to search a robust backbone. 

Overall, our main contributions can be summarized as follows:
\begin{enumerate}[1)]
    \item We propose a plug-in edge-enhanced diverse branch block by revisiting existing re-parameterization technologies. The block can improve the SR performance without extra cost for inference.
    \item We design a novel gradient-variance loss function for edge information preserve. The loss can work with the proposed EDBB to achieve higher restoration quality.
    \item We include block composing, NAS, and loss design into our EFDN framework. And our model achieves a competitive performance while maintaining an extremely lightweight inference.
\end{enumerate}
\section{Related Work}
\subsection{Efficient image super-resolution}
In recent years, convolutional neural networks (CNNs) have greatly promoted the development of low-level computer vision tasks~\cite{ResNet}. In the super-resolution field, Dong \etal~\cite{SRCNN} proposed SRCNN, the earliest CNN-based work which outperforms the traditional methods. However, SRCNN adopts post-upscaling architecture and large convolution layers, which would result in reduced operating efficiency. To remove the unnecessary computational cost,   the authors re-implemented the upscaling module by a deconvolution layer and moved it to the tail part in \cite{FSRCNN}. Since then, plenty of CNN-based SR networks\cite{EDSR,RCAN,HAN,RDN} has been introduced to improve the reconstruction results. Nevertheless, most approaches leverage hundreds of convolution layers and attention mechanisms for higher quality while ignoring the applications under restricted resources. 

To develop the efficient super-resolution in edge devices, Ahn \etal~\cite{CARN} proposed CARN-M, a residual network with a cascading mechanism, which can reduce parameters and computations at the expense of quality reduction. Hui \etal proposed an information distillation network~\cite{IDN} to  explicitly split the intermediate feature to distill and compress the local long and short-path features. Based on IDN, IMDN~\cite{IMDN} is introduced with a more reasonable feature distillation mechanism and effective adaptive cropping strategy. Enlightened by revisiting these distillation mechanisms, Liu \etal~\cite{RFDN} proposed a novel channel splitting strategy that utilizes the convolution layer to implement dimensional change. Furthermore, they devise shallow residual blocks to improve the construction performance while maintaining the parameter scale. With these improvements, they won first place in the AIM 2020 efficient super-resolution challenge~\cite{AIM2020}.

\subsection{Re-parameterization}
Re-parameterization has become an effective technique for efficient neural network design. Lei \etal~\cite{ACB} proposed an asymmetric convolution block (ACB) to strengthen the vanilla convolution by merging three different convolutions. Then in the AIM 2020~\cite{AIM2020}, FIMDN adopted the block into the super-resolution field and outperformed IMDN with fewer parameters. RepVGG~\cite{RepVGG} first added identity mapping and 1$\times$1 convolution into the re-parameterizable structure family, and DBB~\cite{DBB} further enriched the family with sequential convolutions like expanding-and-squeezing convolution. Based on these works, Zhang \etal~\cite{ECB} proposed an edge-oriented convolution block (ECB) to improve the performance of the real-time SR network in mobile devices. 

\begin{figure}[h]
  \centering
  \begin{subfigure}{1.0\linewidth}
  \includegraphics[width=1.0\linewidth]{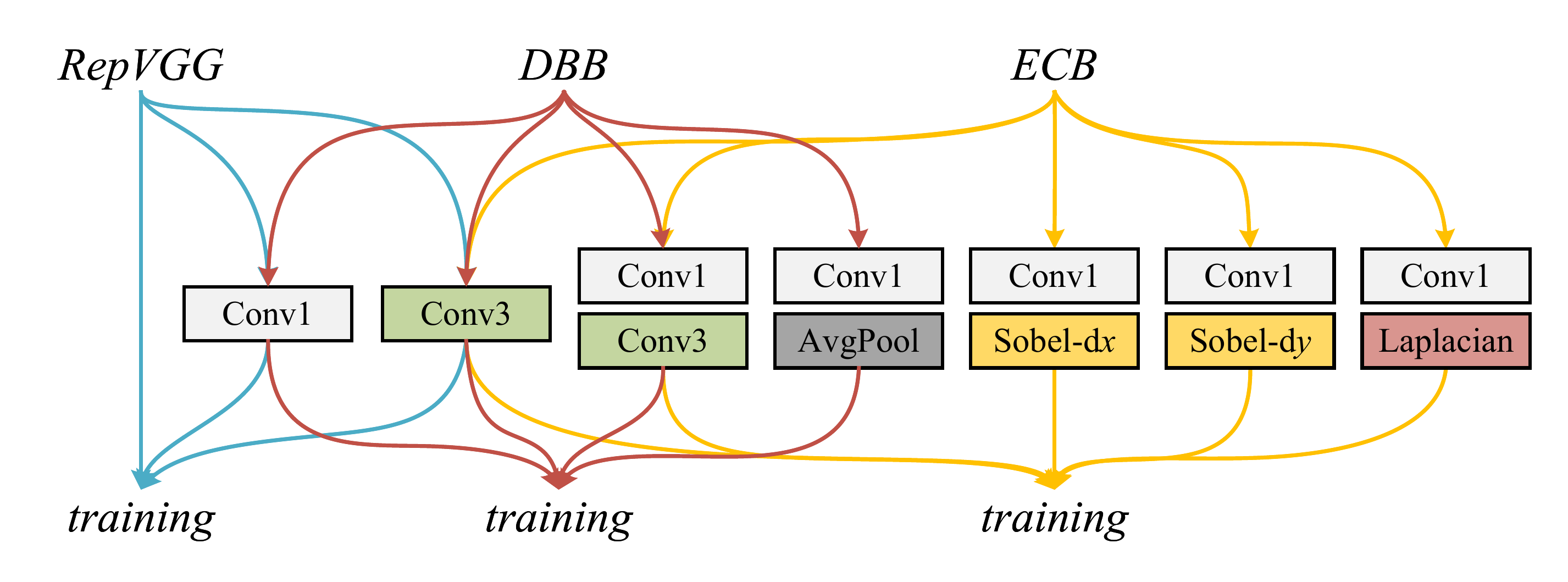}
    \caption{Revisiting re-parameterizable typology.}
    \label{fig:Block-a}
  \end{subfigure}
  \\
  \begin{subfigure}{1.0\linewidth}
  \includegraphics[width=1.0\linewidth]{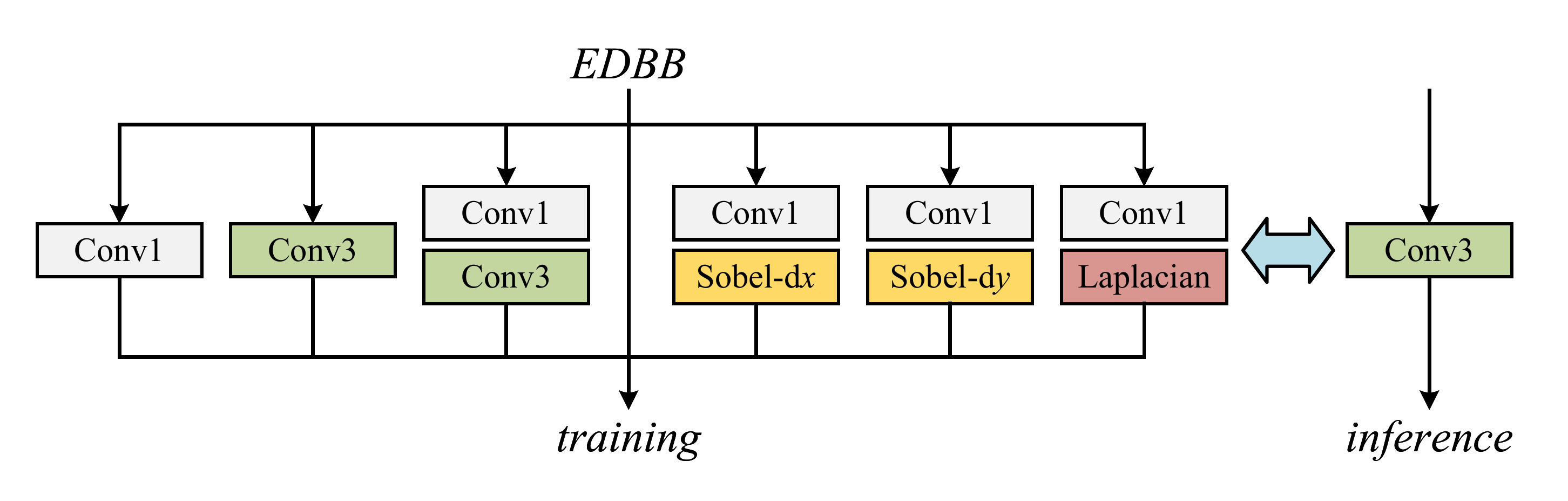}
    \caption{Proposed edge-enhanced diverse branch block.}
    \label{fig:Block-b}
  \end{subfigure}
  \caption{Illustration of re-parameterization method.}
  \label{fig:short}
\end{figure}

\section{Proposed Method}
\label{sec:method}
\subsection{Edge-enhanced diverse branch block}

\begin{figure*}[t]
  \centering
   \includegraphics[width=1.0\linewidth]{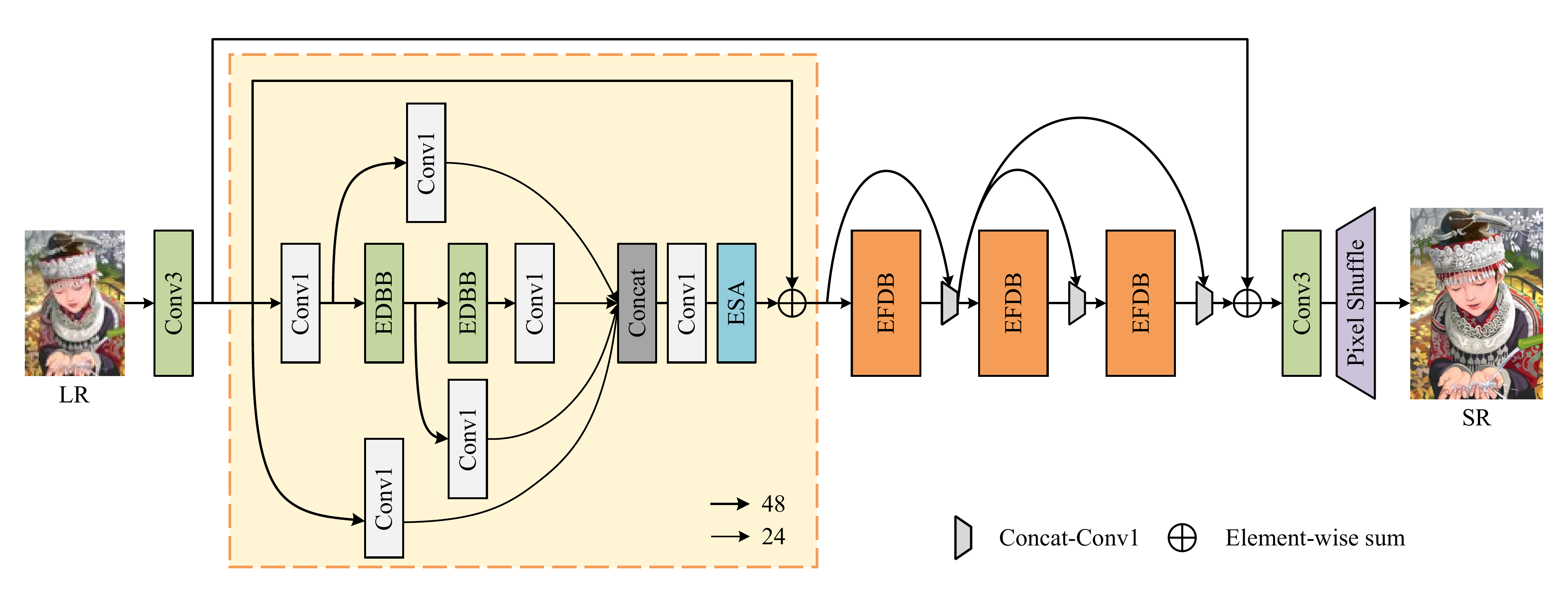}

   \caption{Network architecture of the proposed EFDN.}
   \label{fig:Network}
\end{figure*}

First, we revisit the existing re-parameterizable topology~\cite{RepVGG,ECB,DBB}. As shown in \cref{fig:Block-a}, we present the detail of RepVGG Block, DBB, and ECB. A total of eight different structures have been designed to improve the feature extraction ability of the vanilla convolution in different scenarios. Although the performance may be higher with more re-parameterizable branches, the expensive training cost is unaffordable for straightly integrating these paths. Meanwhile, another problem is that edge and structure information may be attenuated during the merging of parallel branches. 

To address the above concerns, we build a more delicate and effective reparameterization block, namely Edge-enhanced Diverse Branch Block (EDBB), which can extract and preserve high-level structural information for the low-level task. As illustrated in \cref{fig:Block-b}, the EDBB consists of seven branches, summarized in following two categories.

\textbf{Category \uppercase\expandafter{\romannumeral1}: a single convolution.} A normal single convolution operator can be given as:
\begin{equation}
  \sigma (\bm{x}) = \bm{W}\ast\bm{x} + \bm{b},
  \label{conv}
\end{equation}
Given the convolution operator ($\bm{W}$, $\bm{b}$) with a $h$$\times$$w$ kernel and the target convolution ($\bm{W}^{\rm\uppercase\expandafter{\romannumeral1}}$, $\bm{b}^{\rm\uppercase\expandafter{\romannumeral1}}$) with the kernel shape of $H$$\times$$W$ ($h\leq H$,$w\leq W$), the process of assigning reparameterized kernel from zero matrix can be given as:
\begin{subequations} 
  \begin{align}
   \bm{W}^{\rm\uppercase\expandafter{\romannumeral1}} _{:,:,i+\lfloor\frac{H-h}{2}\rfloor,j+\lfloor\frac{W-w}{2}\rfloor} &= \bm{W}_{:,:,i,j},\\
   \bm{b}^{\rm\uppercase\expandafter{\romannumeral1}} &= \bm{b},  \label{rep_1} 
\end{align}
\end{subequations}
While for shortcut operator, it can be treated as a special 1$\times$1 convolution, where $\bm{W}_{:,:,0,0}$ is an identity matrix.

\textbf{Category \uppercase\expandafter{\romannumeral2}: sequential convolutions.} Sequential convolutions are widely applied to further extract the hidden information from the feature maps. In the EDBB, we introduce expanding-and-squeezing convolution and scaled filter convolution to enhance the edge and structure signals. Generally, these re-parameterizable convolution sequences can be expressed by:
\begin{equation}
  \sigma_{2}(\sigma_{1} (\bm{x})) = \bm{W}_{2}\ast(\bm{W}_{1}\ast\bm{x} + \bm{b}_{1}) + \bm{b}_{2},
  \label{seqconv}
\end{equation}
where the $\sigma_{1}$ is the first $D\times C\times1\times 1 $ convolution ($\bm{W}_{1}$, $\bm{b}_{1}$), and $\sigma_{2}$ is the second $C\times D\times K\times K $ convolution ($\bm{W}_{2}$, $\bm{b}_{2}$). To merge them into a $C\times C\times K\times K$ convolution ($\bm{W}^{\rm\uppercase\expandafter{\romannumeral2}}$, $\bm{b}^{\rm\uppercase\expandafter{\romannumeral2}}$), we transform the formula as the form of \cref{conv}. According to \cite{ECB,DBB}, we can obtain the target kernel in the following manner:
\begin{subequations} 
  \begin{align}
   \bm{W}^{\rm\uppercase\expandafter{\romannumeral2}} &= perm(\bm{W}_{1})\ast\bm{W}_{2},\\
   \bm{b}^{\rm\uppercase\expandafter{\romannumeral2}} &= \bm{W}_{2}\ast rep(\bm{b}_1) + \bm{b}_2, 
\end{align}
\end{subequations}
where $perm$ and $rep$ are permuting and broadcasting operations to align weights correspondingly. For $\bm{W}_{1}$, the first two dimensions are exchanged to maintain the same size as $\bm{W}_{2}$.
For $\bm{b}_1$, it is replicated to share the same shape as $\bm{b}_2$.

In the training stage, we train the model with EDBB to obtain more reasonable intermediate features. And then, we transfer EDBB into a vanilla convolution by calculating the sum of re-parameterized parameters. In general, the EDBB leads to quality promotion by utilizing more diverse re-parameterizable branches, and it maintains the running-time inference of vanilla convolution. 


\subsection{Network architecture}

Following IMDN\cite{IMDN} and RFDN\cite{RFDN}, we devise an edge-enhanced information distillation network (EFDN) to reconstruct high-quality SR images with sharp edges and clear structure under restricted resources. As illustrated in \cref{fig:Network}, our EFDN consists of a shallow feature extraction module, multiple edge-enhanced feature distillation blocks (EFDBs), and upscaling module. Specifically, we leverage a single vanilla convolution to generate the initial feature maps. Given the input LR image $I^{LR}$, this information extraction process can be encapsulated as:

\begin{equation}
  F_{0} = E(I^{LR}),
  \label{FE}
\end{equation}
where the $E$ denotes the feature extraction function by a 3$\times$3 convolution, and $F_{0}$ is the extracted feature maps. This coarse feature is then sent to stacked EFDBs for further information refining. In detail, we replace the shallow residual block in \cite{RFDN} with the proposed EDBB to construct our EFDB. Different from IMDN and RFDN utilizing dense distillation connections to process input features progressively, we adopt network-level NAS strategy proposed in DLSR\footnote{\url{https://github.com/DawnHH/DLSR-PyTorch}}~\cite{DLSR} to decide the feature connection paths. The searched structure is shown in the orange dashed box. We denote the proposed EFDB as $H_{n}$, the information distillation procedure can be described as:
\begin{equation}
 \begin{aligned}
 F_{1} &= H_1(F_0),\\
 F_{2} &= H_2(F_1),\\
 F_{3} &= H_3(C_1(F_1,F_2)),\\ 
 F_{4} &= C_3(F_2,H_4(C_2(F_2,F_3)))+F_0,\\
 \end{aligned}
 \label{MD}
\end{equation}
where $C_{n}$ is $n$-th fusion operator consisting of concatenation and 1$\times$1 convolution. $F_n$ is the $n$-th output feature. Finally, the SR images are generated by upscaling module:
\begin{equation}
  I^{SR} = R(F_4),
  \label{UP}
\end{equation}
where $R$ consists of a 3$\times$3 convolution and sub-pixel operation to convert feature maps to images.
\subsection{Edge-enhanced gradient-variance loss}
In previous work\cite{EDSR}, $\mathcal{L}_{1}$ and $\mathcal{L}_{2}$ loss have been in common usage to obtain higher evaluation indicators. The network trained with these loss functions often leads to the loss of structural information. Although the edge-oriented components are added to the EDBB, it is hard to ensure their effectiveness during the complex training procedure of seven parallel branches.
Inspired by the gradient variance (GV) loss~\cite{GV}, we proposed an edge-enhanced gradient-variance (EG) loss, which utilizes the filters of the EDBB to monitor the optimization of the model. In detail, the HR image $I^{HR}$ and SR image $I^{SR}$ are transferred to gray-scale images $G^{HR}$ and $G^{SR}$. We leverage the Sobel and Laplacian filters to compute the gradient maps and then unfold gradient maps into $\frac{HW}{n^2}\times n^2$ patches $G_{x}$, $G_{y}$, $G_{l}$. The $i$-th variance maps can be formulated as:
\begin{equation}
  v_i = \frac{\sum^{n^2}_{j=1}(G_{i,j}-\bar{G}_i)^2}{n^2-1}
  \label{var}
\end{equation}
where $\bar{G}_i$ is the mean value of the $i$-th patch. Thus, we can calculate the variance metrics $v_{x}$,$v_{y}$,$v_{l}$ of HR and SR images, respectively. Referring to GV-loss, we can obtain the gradient variance loss of different filter by:
\begin{equation} 
  \begin{aligned}
  \mathcal{L}_{x} &= \mathds{E}_{I^{SR}}\|v^{HR}_{x}-v^{SR}_{x}\|_2\\
  \mathcal{L}_{y} &= \mathds{E}_{I^{SR}}\|v^{HR}_{y}-v^{SR}_{y}\|_2\\
  \mathcal{L}_{l} &= \mathds{E}_{I^{SR}}\|v^{HR}_{l}-v^{SR}_{l}\|_2
\label{GV}
\end{aligned}
\end{equation}

Besides, we add $\mathcal{L}_{1}$ to accelerate convergence and improve the restoration performance. In order to better optimize the edge-oriented branches of EDBBs and preserve sharp edges for visual effects, we set trade-off coefficients $\lambda_x$, $\lambda_y$, and $\lambda_l$, which are related to the scaled parameters of corresponding branches. In detail, we replace the last convolution layer with EDBB during the pre-training process to obtain the reasonable scaled parameters $s_x$, $s_y$, and $s_l$ of different branches. Then, we determine the $\lambda$ by calculating the normalized weights of $s$, and we transfer the EDBB back into a vanilla convolution for more accessible training. The summative loss function can be expressed by:
\begin{equation}
   \mathcal{L}_{EG} = \mathcal{L}_{1} + \lambda_x\mathcal{L}_{x} + \lambda_y\mathcal{L}_{y} + \lambda_l\mathcal{L}_{l} 
  \label{EG}
\end{equation}

\section{Experiments}
\subsection{Datasets and metrics}
In the training stage, we use DIV2K~\cite{div2k} and Flick2K~\cite{EDSR} (DF2K) to train our models. A total of 3450 images are included to produce high-resolution and bicubic down-sampled image pairs. In the evaluation stage, we utilize four commonly  used benchmark datasets: Set5~\cite{set5}, Set14~\cite{set14}, BSD100~\cite{B100}, and Urban100\cite{Urban100}. In terms of evaluation metrics, peak signal-to-noise ratio (PSNR) and structural similarity (SSIM)~\cite{SSIM} are calculated in the $Y$ channel of $YCbCr$ form to validate the quality of generated images.

\subsection{Implementation details}
To obtain the LR-HR images pairs, we leverage bicubic interpolation to downscale the 2K resolution images. We augment the training datasets by horizontal flips and $90^{\circ}$ rotations. The HR path size and mini-batch size are determined by the training step. The training procedure can be summarized as follows.
\begin{enumerate}[1)]
\item Pre-training the $\times$2 model on DIV2K. The LR patch size is set to 64$\times$64, and the mini-batch size is 64. $\mathcal{L}_{1}$ loss and Adam optimizer~\cite{ADAM} are utilized in optimization. The initial learning rate is defined as $1\times10^{-3}$. Referring to \cite{PAN}, we employ the cosine annealing learning scheme to accelerate convergence.
\item Training models on DF2K. In this step, The LR patch size is 64$\times$64, and the mini-batch size is 32. We use $\mathcal{L}_{EG}$ in \cref{EG} to provide a better visual effect. The initial learning rate is set to $5\times10^{-4}$.
\item Fine-tuning on DF2K. The LR patch size and mini-batch size are 120$\times$120 and 32, respectively. The $\mathcal{L}_2$ loss is chosen to promote the PSNR value. The learning rate is initialized to $1\times10^{-5}$.
\item Reparameterizing and fine-tuning on DIV2K. The LR patch size and mini-batch size are 160$\times$160 and 8, respectively. The $\mathcal{L}_2$ loss is used, and the learning rate starts from $1\times10^{-6}$.
\end{enumerate}
The proposed method is implemented under the PyTorch framework~\cite{Pytorch} with an NVIDIA RTX 3090 GPU.

\subsection{Model analysis}
In this subsection, we investigate model complexity including model size and running-time, and the effectiveness of EDBB and EG loss. 

\begin{figure}[htb]
  \centering
   \includegraphics[width=1.0\linewidth]{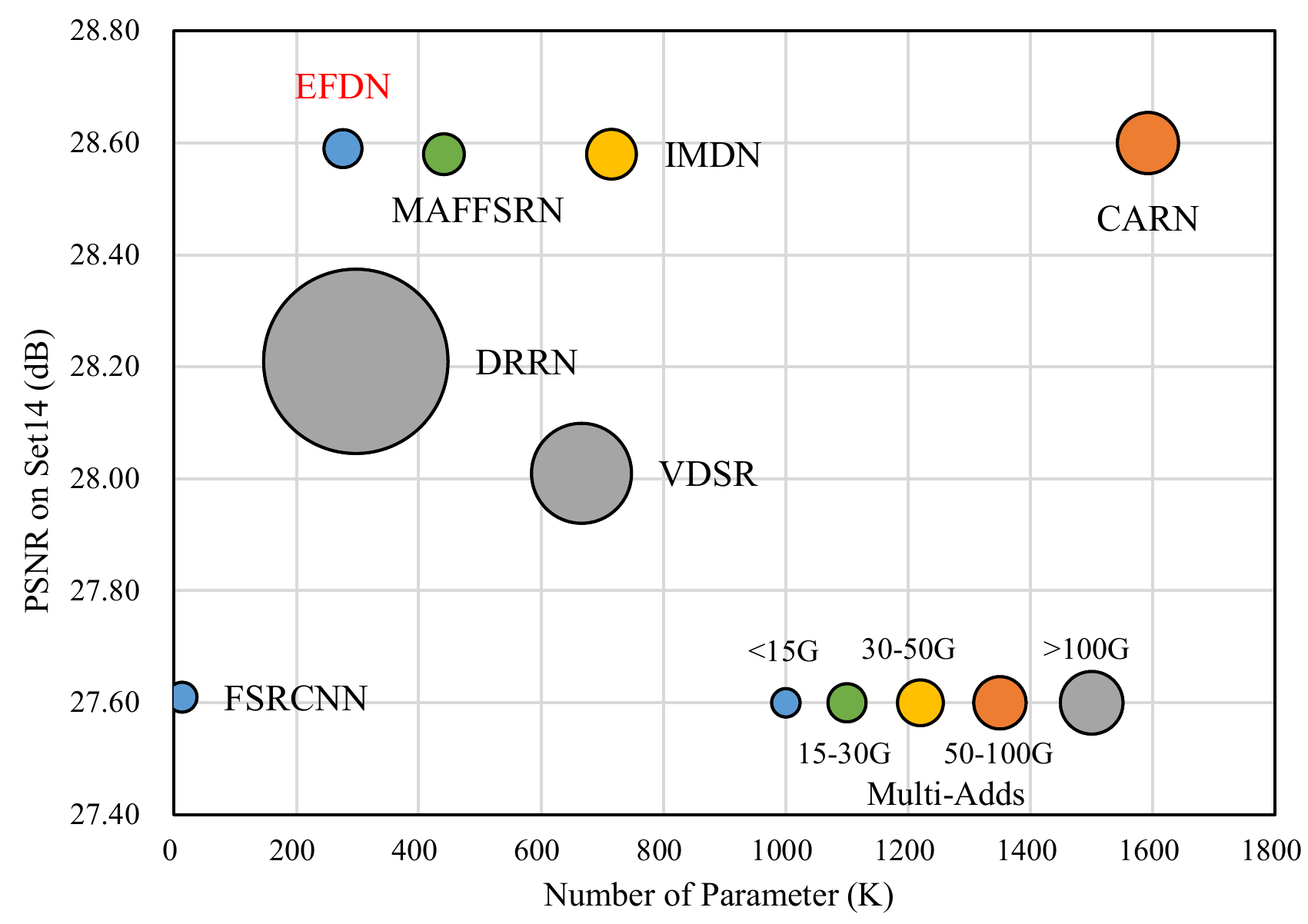}

   \caption{Trade-off between performance and model complexity on Set14~\cite{set14} $\times$4 dataset. The multi-adds operation is calculated with 320$\times$180 input.}
   \label{fig:Model-complex}
\end{figure}

\begin{table}[htb]
  \centering
  \setlength\tabcolsep{2.5pt}
  \begin{tabular}{lcccccc}
    \toprule
    Method & PSNR & Paras. & FLOPs & Act. & Mem. \\
    \midrule
    IMDN~\cite{IMDN}  & \textbf{29.13}/\textbf{28.78} & 893K & 58.53 & 154.14 & \textbf{120}\\     
    RFDN~\cite{RFDN} & 29.04/28.75 & 433K & 27.10 & 112.03 & 200\\
    PAN~\cite{PAN}  & 29.01/28.70 & \textbf{272}K & 32.19 & 270.53 & 311\\   
    EFDN (Ours) & 29.00/28.66 & 276K & \textbf{16.73} & \textbf{111.12} & 168 \\
    \bottomrule
  \end{tabular}
  \caption{Results on method complexity (number of parameters, FLOPs, GPU memory consumption, number of activations).}
  \label{tab:r1}
\end{table}

\subsubsection{Model compexity} 

In \cref{fig:Model-complex}, we provide an overview of our Network's deployment performance. We can find that our EFDN obtains a better trade-off between the SR quality and model size. To evaluate the method complexity of our EFDN precisely, we compare several lightweight networks in \cref{tab:r1}. For fairness, the benchmark in AIM 2020~\cite{AIM2020} is leveraged to evaluate the inference performance. The table shows that our EFDN achieves comparable PSNR on validation and test datasets while consuming fewer resources. In detail, only 16.73G FLOPs are spent during the reconstruction, which is about half of PAN~\cite{PAN} and 29\% of IMDN~\cite{IMDN}.  Moreover, the EFDN also has the minimum activation operations and the second least parameters and memory-consuming among these models. In terms of running time, we compare our approach with IMDN on an NVIDIA RTX 3090 and an RTX 2070-maxQ to evaluate the efficiency in \cref{tab:r2}. Our EFDN is significantly faster than IMDN on different GPUs.

\begin{table}[htb]
  \centering
  \setlength\tabcolsep{7pt}
  \begin{tabular}{lcc}
    \toprule
    Method & Time on 2070-maxQ & Time on 3090 \\
    \midrule
    IMDN~\cite{IMDN}  & 0.158s & 0.092s \\     
    EFDN (Ours) & \textbf{0.089}s & \textbf{0.019}s \\
    \bottomrule
  \end{tabular}
  \caption{Results on running-time.}
  \label{tab:r2}
\end{table}

\begin{table*}[htb]
  \centering
  \small
  \setlength\tabcolsep{3.7pt}
  \caption{Ablation study of re-parameterizable typology and loss function on Set5~\cite{set5}, Set14~\cite{set14}, B100~\cite{B100}, Urban100~\cite{Urban100} ($\times$2). We compare the our EDBB to existing blocks on FSRCNN and test different designs on VDSR. The overall improvement by EDBB and $\mathcal{L}_{EG}$ are valided in FSRCNN, VDSR, and *EFDN. The best/second-best results are \textbf{highlighted} and \underline{underlined}. (FSRCNN and VDSR are re-implemented, and BN layers are discarded during experiments.)}
  \begin{tabular}{llcccccccccc}
    \toprule
      & Block & \makecell[c]{3$\times$3\\ Conv}   &  \makecell[c]{ 1$\times$1 \\Conv}  &Identity & \makecell[c]{Expand-\\  Squeeze}  &\makecell[c]{Scaled Filter}   & Set5 &Set14&B100 &Urban100\\%
    \midrule
     \multirow{7}*{\rotatebox{90}{FSRCNN~\cite{FSRCNN}}}&
     Baseline-$\mathcal{L}_1$ & \checkmark & & & & &37.09/0.9569 &32.75/0.9098 &31.56/0.8913 &30.00/0.9037\\
    & Baseline-$\mathcal{L}_{EG}$ & \checkmark & & & & &37.14/0.9571 &32.77/0.9103 &31.58/0.8917 &30.00/0.9040\\
    & RepVGG~\cite{RepVGG}  & \checkmark  & \checkmark  & \checkmark & & &37.15/0.9571 &32.78/0.9102 &31.59/0.8916 &30.06/0.9045\\
    & DBB~\cite{DBB}  & \checkmark  & \checkmark &  & \checkmark  & Avgpool &37.18/0.9572 &32.77/0.9103 &31.60/0.8918 &30.11/0.9050\\  
    & ECB~\cite{ECB}   & \checkmark  & &  & \checkmark  & Laplacian \& Sobel &37.17/0.9572 &\underline{32.80}/0.9103 &31.59/0.8915 &30.09/0.9044\\
    & EDBB-$\mathcal{L}_1$  & \checkmark & \checkmark & \checkmark & \checkmark & Laplacian \& Sobel & \underline{37.19}/\underline{0.9573} &\underline{32.80}/\underline{0.9104} & \underline{31.61}/\underline{0.8919} & \underline{30.14}/\underline{0.9052}\\
    & EDBB-$\mathcal{L}_{EG}$  & \checkmark & \checkmark & \checkmark & \checkmark & Laplacian \& Sobel & \textbf{37.27}/\textbf{0.9576} &\textbf{32.86}/\textbf{0.9109} & \textbf{31.65}/\textbf{0.8926} & \textbf{30.25}/\textbf{0.9069}\\ 
    \midrule
   \multirow{7}*{\rotatebox{90}{VDSR~\cite{VDSR}}} &Baseline-$\mathcal{L}_1$ & \checkmark & & & & &37.69/0.9593 &33.24/0.9142 &31.99/0.8970 &31.30/0.9198\\
   &Baseline-$\mathcal{L}_{EG}$ & \checkmark & & & & &37.72/0.9595 &33.30/\underline{0.9147} &32.02/\underline{0.8978} &31.40/\underline{0.9215}\\
    & EDBB-$\mathcal{L}_1$  & \checkmark & & \checkmark & \checkmark & Laplacian \& Sobel &37.73/0.9594 &33.26/0.9143 &31.99/0.8968 &31.32/0.9205\\
    & EDBB-$\mathcal{L}_1$  & \checkmark & \checkmark &  & \checkmark & Laplacian \& Sobel &37.73/\underline{0.9596} &\underline{33.33}/0.9145 &32.02/0.8973 &31.38/0.9205\\  
    & EDBB-$\mathcal{L}_1$  & \checkmark & \checkmark & \checkmark & \checkmark & Avgpool & 37.68/0.9593 &33.28/0.9142 &32.00/0.8971 &31.27/0.9190\\
    & EDBB-$\mathcal{L}_1$ & \checkmark & \checkmark & \checkmark & \checkmark & Laplacian \& Sobel & \underline{37.76}/0.9595 & \underline{33.33}/\underline{0.9147} &\underline{32.03}/\underline{0.8975} &\underline{31.41}/0.9207\\ 
    & EDBB-$\mathcal{L}_{EG}$ & \checkmark & \checkmark & \checkmark & \checkmark & Laplacian \& Sobel & \textbf{37.85}/\textbf{0.9600} & \textbf{33.41}/\textbf{0.9158} &\textbf{32.10}/\textbf{0.8987} &\textbf{31.65}/\textbf{0.9237}\\ 
    \midrule
     {\multirow{2}*{\rotatebox{90}{*}}}&
     Baseline-$\mathcal{L}_1$ & \checkmark & & & & & \underline{37.91}/\underline{0.9601} &\underline{33.44}/\underline{0.9168} &\underline{32.12}/\underline{0.8990} & \underline{31.82}/\underline{0.9253}\\
    & EDBB-$\mathcal{L}_{EG}$ & \checkmark & \checkmark & \checkmark & \checkmark & Laplacian \& Sobel & \textbf{38.00}/\textbf{0.9604} &\textbf{33.57}/\textbf{0.9179} &\textbf{32.18}/\textbf{0.8998} &\textbf{32.05}/\textbf{0.9275}\\
    \bottomrule
  \end{tabular}
  \label{tab:r3}
\end{table*}

\begin{table*}[htb]
  \centering
  \small
  \setlength\tabcolsep{2.51pt}
  \caption{Average PSNR/SSIM for scale $\times$2 and $\times$4 on datasets Set5~\cite{set5}, Set14~\cite{set14}, B100~\cite{B100}, Urban100~\cite{Urban100} with bicubic degradation. The parameters and multi-adds are calculated with 1280$\times$720 shape output.The best/second-best results are \textbf{highlighted} and \underline{underlined}.}
  \begin{tabular}{cccccccccc}
    \toprule
             \multirow{2}{*}{Dataset}        & \multirow{2}{*}{Scale}   &Bicubic    &FSRCNN~\cite{FSRCNN}     &VDSR~\cite{VDSR}   &IDN~\cite{IDN}& CARN~\cite{CARN}&IMDN~\cite{IMDN}&PAN~\cite{PAN}&\textbf{EFDN}~(Ours)\\%
             & & Para/MAdds &12K/4.6G &665K/612.6G &553K/31.1G &1592K/90.9G &715K/41.0G &272K/28.2G &276K/14.7G \\
    \midrule
    \multirow{2}{*}{Set5}     & $\times$2 &33.66/0.9299 &37.00/0.9558 &37.53/0.9587 &37.83/0.9600           &37.76/0.9590         &\textbf{38.00}/\textbf{0.9605}        &\textbf{38.00}/\textbf{0.9605} &\textbf{38.00}/\underline{0.9604}\\ 
                              & $\times$4 &28.42/0.8104 & 31.35/0.8838  &31.82/0.8903    &32.13/0.8937         &\textbf{32.21}/\textbf{0.8948}         &32.13/\textbf{0.8948}        &32.13/\textbf{0.8948} & {32.08}/{0.8931} \\
    \midrule
    \multirow{2}{*}{Set14}    & $\times$2 &30.24/0.8688 &32.63/0.9088  &33.03/0.9124 &33.30/0.9148                &33.52/0.9166       &\textbf{33.63}/0.9177 &\underline{33.59}/\textbf{0.9181}  & 33.57/\underline{0.9179}\\ 
                              & $\times$4 &26.00/0.7027&   27.61/0.7550 & 28.01/0.7674    &28.25/0.7730         &\textbf{28.60}/0.7806         &\underline{28.58}/\underline{0.7811}        &\textbf{28.60}/\textbf{0.7822}& \underline{28.58}/0.7809\\
    \midrule
    \multirow{2}{*}{B100}     & $\times$2 &29.56/0.8403 &31.53/0.8920 &31.90/0.8960 &32.08/0.8985         &32.09/0.8978        &\textbf{32.19}/0.8996        &\underline{32.18}/\underline{0.8997} &\underline{32.18}/\textbf{0.8998}\\ 
                              & $\times$4 &25.96/0.6675 &26.98/0.7150  &27.29/0.7251 &27.41/0.7297    &\underline{27.58}/0.7349         &27.56/0.7353         &\textbf{27.59}/\textbf{0.7363}       & 27.56/\underline{0.7354} \\
    \midrule
    \multirow{2}{*}{Urban100} & $\times$2 &26.88/0.8403&29.88/0.9020 &30.76/0.9140    &31.27/0.9196         &31.92/0.9256         &\textbf{32.17}/\textbf{0.9283}        &32.01/0.9273 &\underline{32.05}/\underline{0.9275}\\ 
                              & $\times$4 &23.14/0.6577&   24.62/0.7280 &25.18/0.7524    &25.41/0.7632         &\underline{26.07}/0.7837         &26.04/\underline{0.7838}        &\textbf{26.11}/\textbf{0.7854} &26.00/0.7815 \\
    \bottomrule
  \end{tabular}
  \label{tab:r4}
\end{table*}

\subsubsection{Ablation studies of EDBB and EG loss}
To evaluate the effectiveness of the proposed EDBB, we first compare it to other re-parameterization approaches on FSRCNN~\cite{FSRCNN} in \cref{tab:r3}. We can observe that all four blocks can improve the PSNR/SSIM values, while our EDBB leads to higher performance improvement (PSNR: \textbf{+0.05$\thicksim$0.14dB}, SSIM: \textbf{+0.0004$\thicksim$0.0015} on FSRCNN). We also examine the EDBB on the deeper  VDSR~\cite{VDSR} framework by removing or replacing re-parameterizable components. The results in \cref{tab:r3} suggest that any branch change may lead to a quality drop. Overall, we employ EDBB as the core feature extractor in the EFDN backbone to improve the SR performance.

Additionally, we assess the impact of loss functions on the final resolving quality. In detail, we leverage $\mathcal{L}_{1}$ and $\mathcal{L}_{EG}$ onto the baseline and EDBB model, respectively. As listed in \cref{tab:r3}, the baseline trained by $\mathcal{L}_{EG}$ brings a slight improvement on PSNR but a performance boost on SSIM. For instance, the PSNR and SSIM of Urban100 are developed by 0.03dB and 0.0008 on VDSR, respectively. For models equipping EDBB, the benefits of using EG-loss are more significant. 
The combination method of EDBB and $\mathcal{L}_{EG}$ surpasses baseline by a large margin with 0.2dB improvement on three frameworks. Specifically, PSNR/SSIM results of our EFDN increased by 0.23dB/0.002 on the Urban100 testset. Moreover, we can infer from the advance of the SSIM index that the proposed $\mathcal{L}_{EG}$ helps the structure reconstruction by introducing edge-enhanced filters to calibrate EDBB training.

\subsection{Comparison with state-of-the-arts}

\begin{figure*}[!t]
  \vspace{-0.3 cm}
  \setlength\tabcolsep{2.0pt}
  \centering
  \footnotesize
  \begin{tabular}{ccccc}
  \multirow{-6.85}{*}{\includegraphics[width=.34\linewidth, height=4.8cm]{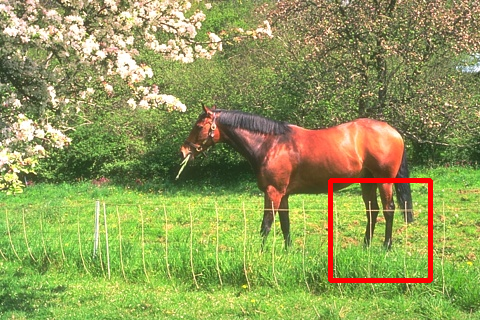}} 
   & \includegraphics[width=.13\linewidth, height=2.2cm]{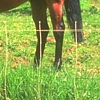} & \includegraphics[width=.13\linewidth, height=2.2cm]{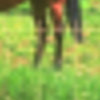}& \includegraphics[width=.13\linewidth, height=2.2cm]{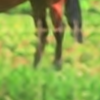}& \includegraphics[width=.13\linewidth, height=2.2cm]{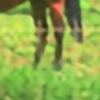} \\
   & HR & Bicubic & SRCNN & FSRCNN\\
   & \includegraphics[width=.13\linewidth, height=2.2cm]{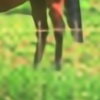} & \includegraphics[width=.13\linewidth, height=2.2cm]{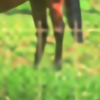} & \includegraphics[width=.13\linewidth, height=2.2cm]{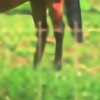}& \includegraphics[width=.13\linewidth, height=2.2cm]{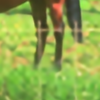} \\
   \emph{291000} from B100~\cite{B100} & VDSR& CARN & IMDN & EFDN~(Ours) \\
   \multirow{-6.85}{*}{\includegraphics[width=.34\linewidth, height=4.8cm]{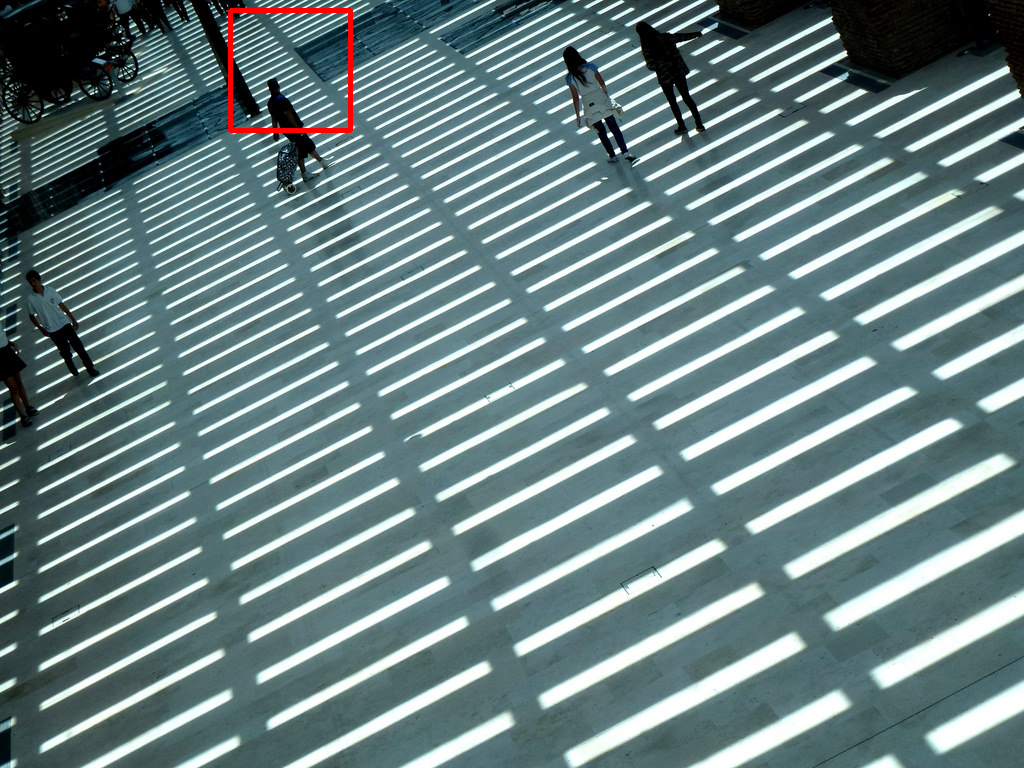}} 
   & \includegraphics[width=.13\linewidth, height=2.2cm]{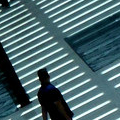} & \includegraphics[width=.13\linewidth, height=2.2cm]{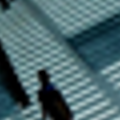}& \includegraphics[width=.13\linewidth, height=2.2cm]{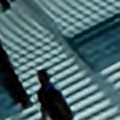}& \includegraphics[width=.13\linewidth, height=2.2cm]{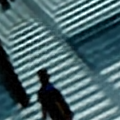} \\
   & HR & Bicubic & SRCNN & FSRCNN\\
   & \includegraphics[width=.13\linewidth, height=2.2cm]{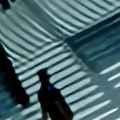} & \includegraphics[width=.13\linewidth, height=2.2cm]{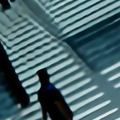} & \includegraphics[width=.13\linewidth, height=2.2cm]{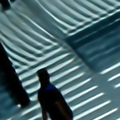}& \includegraphics[width=.13\linewidth, height=2.2cm]{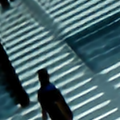} \\
   \emph{img\_093} from Urban100\cite{Urban100} & VDSR& CARN & IMDN & EFDN~(Ours) \\
   \multirow{-6.85}{*}{\includegraphics[width=.34\linewidth, height=4.8cm]{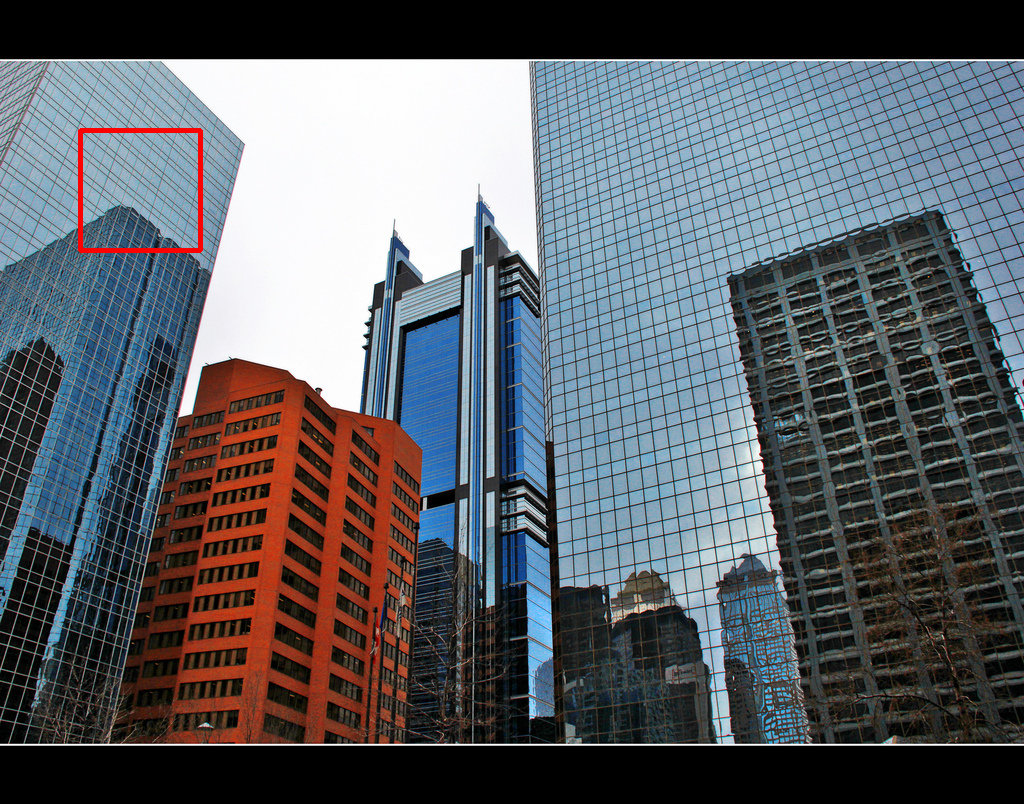}} 
   & \includegraphics[width=.13\linewidth, height=2.2cm]{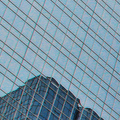} & \includegraphics[width=.13\linewidth, height=2.2cm]{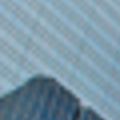}& \includegraphics[width=.13\linewidth, height=2.2cm]{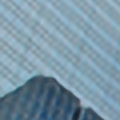}& \includegraphics[width=.13\linewidth, height=2.2cm]{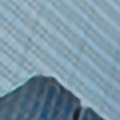} \\
   & HR & Bicubic & SRCNN & FSRCNN\\
   & \includegraphics[width=.13\linewidth, height=2.2cm]{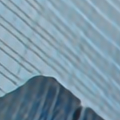} & \includegraphics[width=.13\linewidth, height=2.2cm]{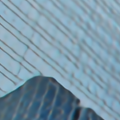} & \includegraphics[width=.13\linewidth, height=2.2cm]{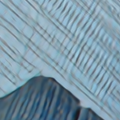}& \includegraphics[width=.13\linewidth, height=2.2cm]{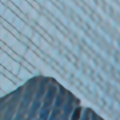} \\
   \emph{img\_099} from Urban100~\cite{Urban100} & VDSR& CARN & IMDN & EFDN~(Ours) \\
\end{tabular}
  \caption{Qualitative results of lightweight SR models with an upscaling factor $\times$4.}
  \label{tab:Bi_fig}
  \end{figure*}
  
We compare our EFDN with several state-of-the-arts lightweight SR methods~\cite{SRCNN,FSRCNN,VDSR,CARN,IDN,IMDN,PAN} on $\times$2 and $\times$4 tasks in \cref{tab:r4}. We use PSNR/SSIM as well as the numbers of parameters and Multi-adds to show model efficiency. It can be found that our method achieves comparable performance to the IMDN and CARN while using fewer parameters and computations. Specifically, the parameter number of EFDN is only 17\% of CARN, and 38\% of IMDN. For $\times$4 task, the muti-adds operands used in our EFDN are far less than these methods, which is 14.7G, about 53\% in PAN and 16\% in CARN. In general, our EFDN is the most lightweight model to maintain the fidelity performance.  

Apart from numerical results, we also show the visual comparison in \cref{tab:Bi_fig}. From the patch of \emph{291000}, we can observe that our EFDN achieves similar reconstruction results with these high consumption models. In \emph{img\_093} and \emph{img\_099} from Urban100~\cite{Urban100}, EFDN surpasses other methods by better quality and less deformation on structural details such as shadows and windows.

\subsection{Challenge results} 
We have participated in NTIRE 2022 Efficient Super-Resolution Challenge~\cite{li2022ntire}. This competition aims to devise a practical SR method that can maintain the PSNR value of IMDN~\cite{IMDN} on DIV2K~\cite{div2k} validation with less resource consumption. Among the final 35 valid submissions, our EFDN ranks 9th in the running time track, 5th in the model complexity track, and 7th in the overall performance track.

\section{Conclusion}  

In this paper, we propose an edge-enhanced feature distillation network for lightweight and accurate super-resolution. We devise an edge-enhanced diverse branch block, which employs more effective re-parameterizable paths to achieve better extraction capability. Furthermore, we design an edge-enhancing loss to maximize the effectiveness of the EDBB. By introducing these strategies, the model size is significantly and steadily reduced while maintaining a commendable SR performance. Numerous experiments have shown the efficiency of the proposed strengthening approaches. 

{\small
\bibliographystyle{ieee_fullname}
\bibliography{egbib}

\begin{thebibliography}{10}\itemsep=-1pt

\bibitem{GV}
Lusine Abrahamyan, Anh~Minh Truong, Wilfried Philips, and Nikos Deligiannis.
\newblock Gradient variance loss for structure-enhanced image super-resolution.
\newblock {\em arXiv preprint arXiv:2202.00997}, 2022.

\bibitem{div2k}
Eirikur Agustsson and Radu Timofte.
\newblock Ntire 2017 challenge on single image super-resolution: Dataset and
  study.
\newblock In {\em CVPRW}, pages 126--135, Honolulu, HI, USA, 2017.

\bibitem{CARN}
Namhyuk Ahn, Byungkon Kang, and Kyung-Ah Sohn.
\newblock Fast, accurate, and lightweight super-resolution with cascading
  residual network.
\newblock In {\em ECCV}, pages 252--268, 2018.

\bibitem{set5}
Marco Bevilacqua, Aline Roumy, Christine Guillemot, and Marie{-}Line
  Alberi{-}Morel.
\newblock Low-complexity single-image super-resolution based on nonnegative
  neighbor embedding.
\newblock In {\em BMVC}, pages 1--10, Surrey, UK, 2012.

\bibitem{chu2021fast}
Xiangxiang Chu, Bo Zhang, Hailong Ma, Ruijun Xu, and Qingyuan Li.
\newblock Fast, accurate and lightweight super-resolution with neural
  architecture search.
\newblock In {\em ICPR}, pages 59--64. IEEE, 2021.

\bibitem{ACB}
Xiaohan Ding, Yuchen Guo, Guiguang Ding, and Jungong Han.
\newblock Acnet: Strengthening the kernel skeletons for powerful cnn via
  asymmetric convolution blocks.
\newblock In {\em ICCV}, pages 1911--1920, Seoul, Korea (South), 2019.

\bibitem{DBB}
Xiaohan Ding, Xiangyu Zhang, Jungong Han, and Guiguang Ding.
\newblock Diverse branch block: Building a convolution as an inception-like
  unit.
\newblock In {\em CVPR}, pages 10886--10895, 2021.

\bibitem{RepVGG}
Xiaohan Ding, Xiangyu Zhang, Ningning Ma, Jungong Han, Guiguang Ding, and Jian
  Sun.
\newblock Repvgg: Making vgg-style convnets great again.
\newblock In {\em CVPR}, pages 13733--13742, 2021.

\bibitem{SRCNN}
Chao Dong, Chen~Change Loy, Kaiming He, and Xiaoou Tang.
\newblock Image super-resolution using deep convolutional networks.
\newblock {\em IEEE TPAMI}, 38(2):295--307, 2016.

\bibitem{FSRCNN}
Chao Dong, Chen~Change Loy, and Xiaoou Tang.
\newblock Accelerating the super-resolution convolutional neural network.
\newblock In {\em ECCV}, pages 391--407, Amsterdam, The Netherlands, 2016.

\bibitem{ResNet}
Kaiming He, Xiangyu Zhang, Shaoqing Ren, and Jian Sun.
\newblock Deep residual learning for image recognition.
\newblock In {\em CVPR}, pages 770--778, Las Vegas, NV, USA, 2016.

\bibitem{DLSR}
Han Huang, Li Shen, Chaoyang He, Weisheng Dong, Haozhi Huang, and Guangming
  Shi.
\newblock Lightweight image super-resolution with hierarchical and
  differentiable neural architecture search.
\newblock {\em arXiv preprint arXiv:2105.03939}, 2021.

\bibitem{Urban100}
Jia-Bin Huang, Abhishek Singh, and Narendra Ahuja.
\newblock Single image super-resolution from transformed self-exemplars.
\newblock In {\em CVPR}, pages 5197--5206, Boston, MA, USA, 2015.

\bibitem{IMDN}
Zheng Hui, Xinbo Gao, Yunchu Yang, and Xiumei Wang.
\newblock Lightweight image super-resolution with information
  multi-distillation network.
\newblock In {\em ACM MM}, pages 2024--2032, Nice, France, 2019.

\bibitem{IDN}
Zheng Hui, Xiumei Wang, and Xinbo Gao.
\newblock Fast and accurate single image super-resolution via information
  distillation network.
\newblock In {\em CVPR}, pages 723--731, 2018.

\bibitem{VDSR}
Jiwon Kim, Jung~Kwon Lee, and Kyoung~Mu Lee.
\newblock Accurate image super-resolution using very deep convolutional
  networks.
\newblock In {\em CVPR}, pages 1646--1654, Las Vegas, NV, USA, 2016.

\bibitem{DRCN}
Jiwon Kim, Jung~Kwon Lee, and Kyoung~Mu Lee.
\newblock Deeply-recursive convolutional network for image super-resolution.
\newblock In {\em CVPR}, pages 1637--1645, Las Vegas, NV, USA, 2016.

\bibitem{ADAM}
Diederik~P. Kingma and Jimmy Ba.
\newblock Adam: {A} method for stochastic optimization.
\newblock In {\em ICLR}, San Diego, CA, USA, 2015.

\bibitem{li2019learning}
Yawei Li, Shuhang Gu, Luc~Van Gool, and Radu Timofte.
\newblock Learning filter basis for convolutional neural network compression.
\newblock In {\em ICCV}, 2019.

\bibitem{li2020dhp}
Yawei Li, Shuhang Gu, Kai Zhang, Luc~Van Gool, and Radu Timofte.
\newblock Dhp: Differentiable meta pruning via hypernetworks.
\newblock In {\em ECCV}, pages 608--624. Springer, 2020.

\bibitem{li2021heterogeneity}
Yawei Li, Wen Li, Martin Danelljan, Kai Zhang, Shuhang Gu, Luc Van~Gool, and
  Radu Timofte.
\newblock The heterogeneity hypothesis: Finding layer-wise differentiated
  network architectures.
\newblock In {\em CVPR}, pages 2144--2153, 2021.

\bibitem{li2022ntire}
Yawei Li, Kai Zhang, Luc~Van Gool, Radu Timofte, et~al.
\newblock Ntire 2022 challenge on efficient super-resolution: Methods and
  results.
\newblock In {\em CVPRW}, 2022.

\bibitem{EDSR}
Bee Lim, Sanghyun Son, Heewon Kim, Seungjun Nah, and Kyoung Mu~Lee.
\newblock Enhanced deep residual networks for single image super-resolution.
\newblock In {\em CVPRW}, pages 136--144, 2017.

\bibitem{RFDN}
Jie Liu, Jie Tang, and Gangshan Wu.
\newblock Residual feature distillation network for lightweight image
  super-resolution.
\newblock In {\em ECCVW}, volume 12537 of {\em Lecture Notes in Computer
  Science}, pages 41--55, Glasgow, UK, 2020. Springer.

\bibitem{B100}
David Martin, Charless Fowlkes, Doron Tal, and Jitendra Malik.
\newblock A database of human segmented natural images and its application to
  evaluating segmentation algorithms and measuring ecological statistics.
\newblock In {\em ICCV}, volume~2, pages 416--423, Vancouver, British Columbia,
  Canada, 2001.

\bibitem{HAN}
Ben Niu, Weilei Wen, Wenqi Ren, Xiangde Zhang, Lianping Yang, Shuzhen Wang,
  Kaihao Zhang, Xiaochun Cao, and Haifeng Shen.
\newblock Single image super-resolution via a holistic attention network.
\newblock In {\em ECCV}, pages 191--207, Glasgow, UK, 2020.

\bibitem{Pytorch}
Adam Paszke, Sam Gross, Francisco Massa, Adam Lerer, James Bradbury, Gregory
  Chanan, Trevor Killeen, Zeming Lin, Natalia Gimelshein, Luca Antiga, et~al.
\newblock Pytorch: An imperative style, high-performance deep learning library.
\newblock {\em Advances in neural information processing systems}, 32, 2019.

\bibitem{song2020efficient}
Dehua Song, Chang Xu, Xu Jia, Yiyi Chen, Chunjing Xu, and Yunhe Wang.
\newblock Efficient residual dense block search for image super-resolution.
\newblock In {\em AAAI}, volume~34, pages 12007--12014, 2020.

\bibitem{DRRN}
Ying Tai, Jian Yang, and Xiaoming Liu.
\newblock Image super-resolution via deep recursive residual network.
\newblock In {\em CVPR}, pages 3147--3155, 2017.

\bibitem{ANR}
Radu Timofte, Vincent De~Smet, and Luc Van~Gool.
\newblock Anchored neighborhood regression for fast example-based
  super-resolution.
\newblock In {\em ICCV}, pages 1920--1927, 2013.

\bibitem{A+}
Radu Timofte, Vincent De~Smet, and Luc Van~Gool.
\newblock A+: Adjusted anchored neighborhood regression for fast
  super-resolution.
\newblock In {\em ACCV}, pages 111--126. Springer, 2014.

\bibitem{SSIM}
Zhou Wang, Alan~C Bovik, Hamid~R Sheikh, and Eero~P Simoncelli.
\newblock Image quality assessment: from error visibility to structural
  similarity.
\newblock {\em IEEE TIP}, 13(4):600--612, 2004.

\bibitem{wu2021trilevel}
Yan Wu, Zhiwu Huang, Suryansh Kumar, Rhea~Sanjay Sukthanker, Radu Timofte, and
  Luc Van~Gool.
\newblock Trilevel neural architecture search for efficient single image
  super-resolution.
\newblock {\em arXiv preprint arXiv:2101.06658}, 2021.

\bibitem{set14}
Roman Zeyde, Michael Elad, and Matan Protter.
\newblock On single image scale-up using sparse-representations.
\newblock In {\em 7th International Conference on Curves and Surfaces}, volume
  6920, pages 711--730, Avignon, France, 2010.

\bibitem{AIM2020}
Kai Zhang, Martin Danelljan, Yawei Li, Radu Timofte, Jie Liu, Jie Tang,
  Gangshan Wu, Yu Zhu, Xiangyu He, Wenjie Xu, et~al.
\newblock Aim 2020 challenge on efficient super-resolution: Methods and
  results.
\newblock In {\em ECCVW}, pages 5--40. Springer, 2020.

\bibitem{Edge-interp}
Lei Zhang and Xiaolin Wu.
\newblock An edge-guided image interpolation algorithm via directional
  filtering and data fusion.
\newblock {\em IEEE TIP}, 15(8):2226--2238, 2006.

\bibitem{ECB}
Xindong Zhang, Hui Zeng, and Lei Zhang.
\newblock Edge-oriented convolution block for real-time super resolution on
  mobile devices.
\newblock In {\em ACM MM}, pages 4034--4043, 2021.

\bibitem{RCAN}
Yulun Zhang, Kunpeng Li, Kai Li, Lichen Wang, Bineng Zhong, and Yun Fu.
\newblock Image super-resolution using very deep residual channel attention
  networks.
\newblock In {\em ECCV}, pages 286--301, Munich, Germany, 2018.

\bibitem{RDN}
Yulun Zhang, Yapeng Tian, Yu Kong, Bineng Zhong, and Yun Fu.
\newblock Residual dense network for image super-resolution.
\newblock In {\em CVPR}, pages 2472--2481, Salt Lake City, UT, USA, 2018.

\bibitem{PAN}
Hengyuan Zhao, Xiangtao Kong, Jingwen He, Yu Qiao, and Chao Dong.
\newblock Efficient image super-resolution using pixel attention.
\newblock In {\em ECCVW}, pages 56--72. Springer, 2020.

\end{thebibliography}
}

\end{document}